\documentclass[10pt,twocolumn,letterpaper]{article}

\usepackage{cvpr}
\usepackage{times}
\usepackage{epsfig}
\usepackage{graphicx}
\graphicspath{{figs/}}
\usepackage{amsmath}
\usepackage{amssymb}
\usepackage{subfigure}
\usepackage{booktabs}
\usepackage{multirow}
\usepackage{balance}
\usepackage{enumitem}
\usepackage{float}
\usepackage{lipsum}
\usepackage{rotating}
\usepackage[utf8]{inputenc}
\usepackage{bbm}
\usepackage{caption}
\usepackage{soul}
\usepackage{pifont}
\usepackage{color}
\definecolor{gray}{rgb}{0.80, 0.80, 0.80}
\definecolor{orange}{rgb}{1,0.9,0.65}
\definecolor{gr}{rgb}{0.9,1,0.6}
\definecolor{bl}{rgb}{0.9,0.8,1}
\definecolor{bg}{rgb}{0.8,0.9,1}
\definecolor{orr}{rgb}{1,0.92,0.75}
\definecolor{grr}{rgb}{0.8,1,0.5}
\definecolor{blr}{rgb}{0.97,0.8,1}
\definecolor{bgr}{rgb}{0.7,0.9,1}
\definecolor{gry}{rgb}{0.92,0.92,0.92}

\definecolor{cadmiumgreen}{rgb}{0.0, 0.42, 0.24}
\definecolor{cornellred}{rgb}{0.7, 0.11, 0.11}
\definecolor{cornflowerblue}{rgb}{0.39, 0.58, 0.93}

\usepackage{array}
\usepackage{multirow}
\usepackage{colortbl}

\usepackage[pagebackref=true,breaklinks=true,letterpaper=true,colorlinks,bookmarks=false]{hyperref}

\newcommand{\cmark}{{\ding{51}}}%

\newcolumntype{L}[1]{>{\raggedright\let\newline\\\arraybackslash\hspace{0pt}}m{#1}}
\newcolumntype{C}[1]{>{\centering\let\newline\\\arraybackslash\hspace{0pt}}m{#1}}
\newcolumntype{R}[1]{>{\raggedleft\let\newline\\\arraybackslash\hspace{0pt}}m{#1}}

\cvprfinalcopy 

\def\cvprPaperID{1284} 

\ifcvprfinal\fi
\begin{document}

\title{\vspace{-0.5cm}MovieQA: Understanding Stories in Movies through Question-Answering}

\author{
Makarand Tapaswi$^1$,\hspace{1.4cm}Yukun Zhu$^3$,\hspace{1.5cm}Rainer Stiefelhagen$^1$\\
Antonio Torralba$^2$,\hspace{1.0cm}Raquel Urtasun$^3$,\hspace{1.5cm}Sanja Fidler$^3$
\vspace{0.1cm}\\
$^1$Karlsruhe Institute of Technology,
$^2$Massachusetts Institute of Technology,
$^3$University of Toronto
\\
\texttt{\footnotesize
\{tapaswi,rainer.stiefelhagen\}@kit.edu,
torralba@csail.mit.edu,
\{yukun,urtasun,fidler\}@cs.toronto.edu
}
\\
{\normalsize \url{http://movieqa.cs.toronto.edu}}
}


\twocolumn[{%
\renewcommand\twocolumn[1][]{#1}%
\maketitle
\vspace*{-0.7cm}
\centering
  \includegraphics[width=\linewidth,trim=0 36 0 91,clip]{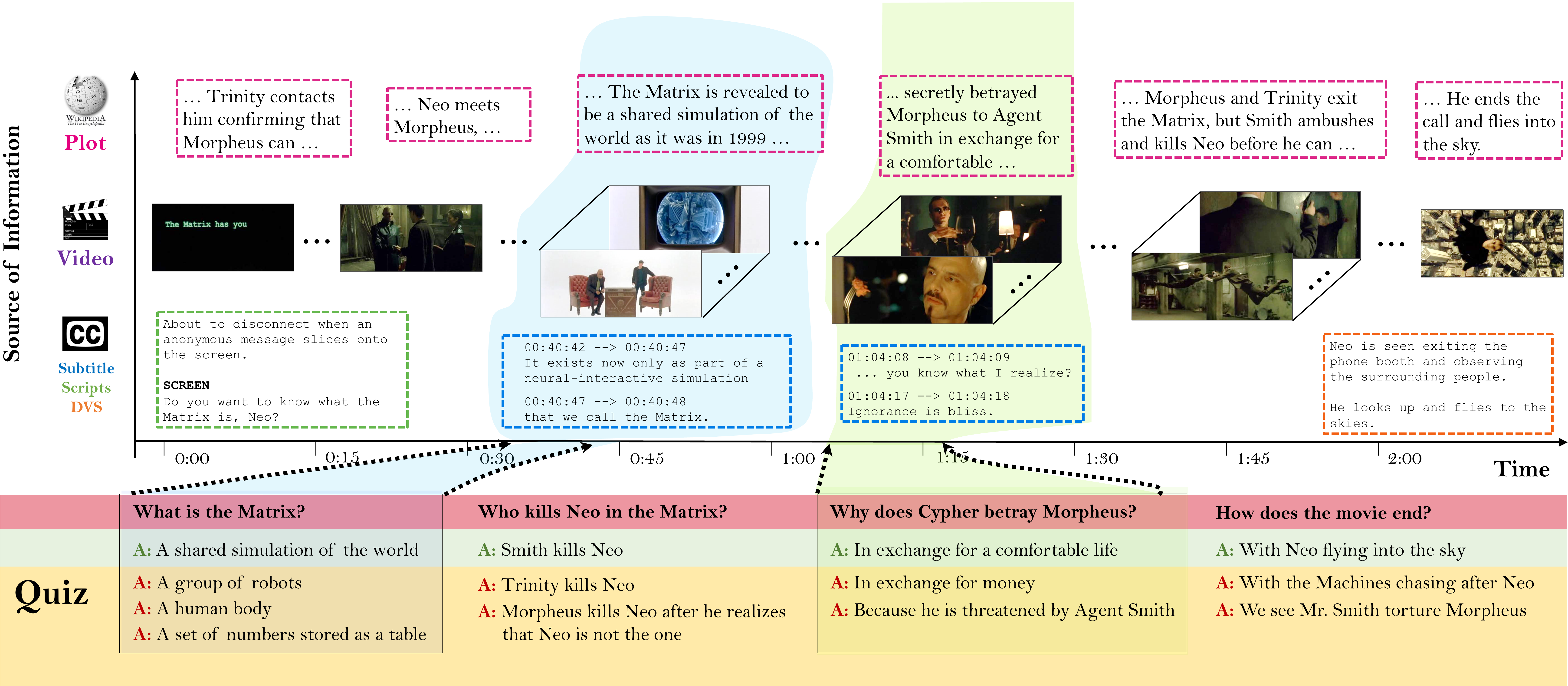}
\vspace*{-0.7cm}
\captionof{figure}{\small Our MovieQA dataset contains 14,944 questions about 408 movies.
It contains multiple sources of information: plots, subtitles, video clips, scripts, and DVS transcriptions.
In this figure we show example QAs from \emph{The Matrix} and localize them in the timeline.}
\label{fig:frontpage}
\vspace*{0.5cm}
}]

\begin{abstract}
We introduce the MovieQA dataset which aims to evaluate automatic story comprehension from both video and text.
The dataset consists of 14,944 questions about 408 movies with high semantic diversity.
The questions range from simpler ``Who'' did ``What'' to ``Whom'', to ``Why'' and ``How'' certain events occurred.
Each question comes with a set of five possible answers; a correct one and four deceiving answers provided by human annotators.
Our dataset is unique in that it contains multiple sources of information -- video clips, plots, subtitles, scripts, and DVS~\cite{Rohrbach15}.
We analyze our data through various statistics and methods.
We further extend existing QA techniques to show that question-answering with such open-ended semantics is hard.
We make this data set public along with an evaluation benchmark to encourage inspiring work in this challenging domain.
\end{abstract}

\vspace{-2mm}
\vspace{-1mm}
\section{Introduction}
\label{sec:intro}
\vspace{-1mm}

\begin{figure*}[t]
\vspace{-3mm}
\includegraphics[height=0.161\linewidth]{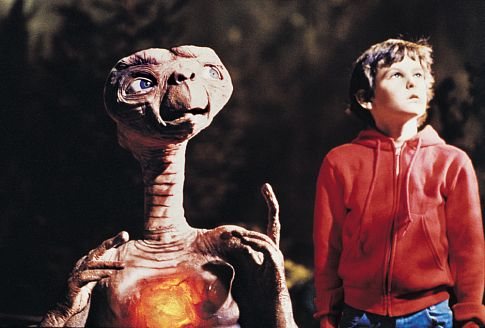}
\includegraphics[height=0.161\linewidth]{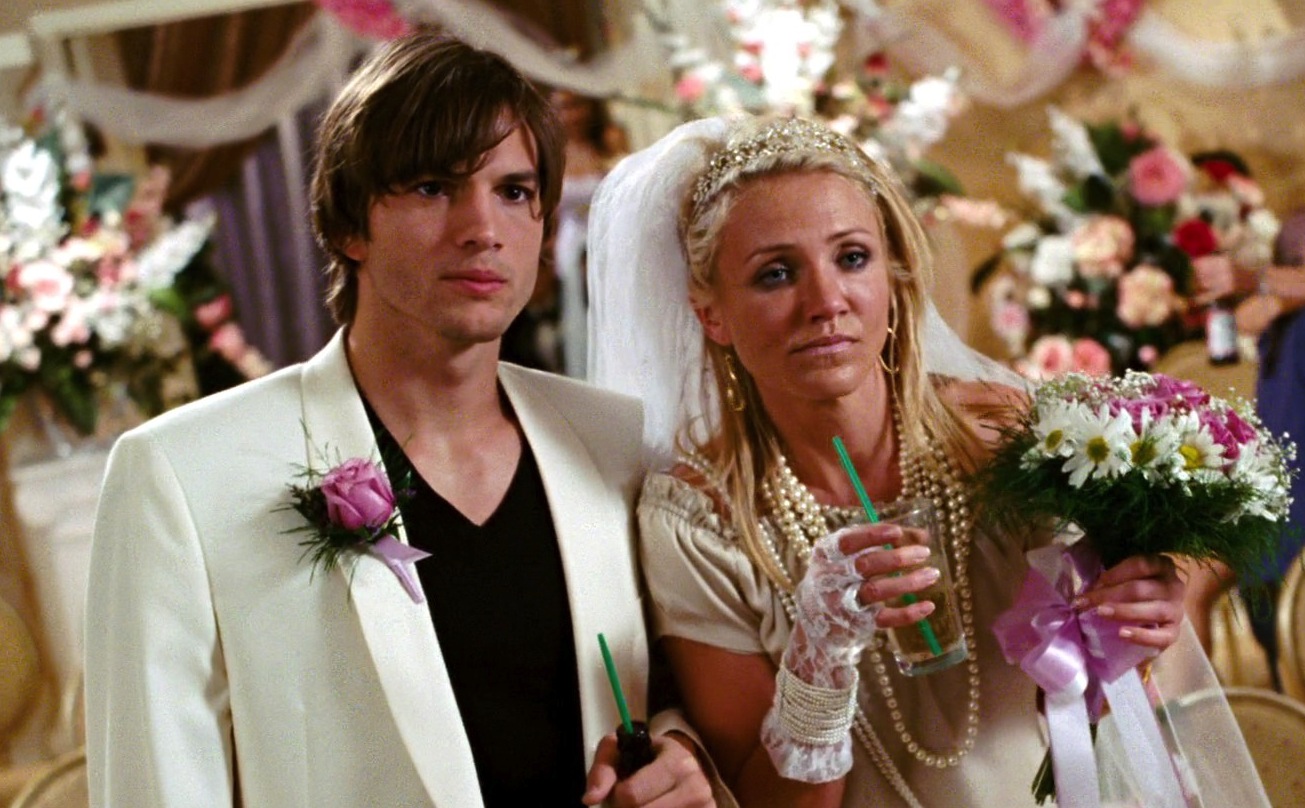}
\includegraphics[height=0.161\linewidth]{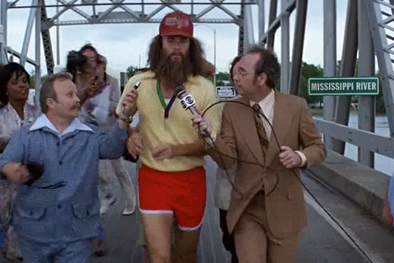}
\includegraphics[height=0.161\linewidth]{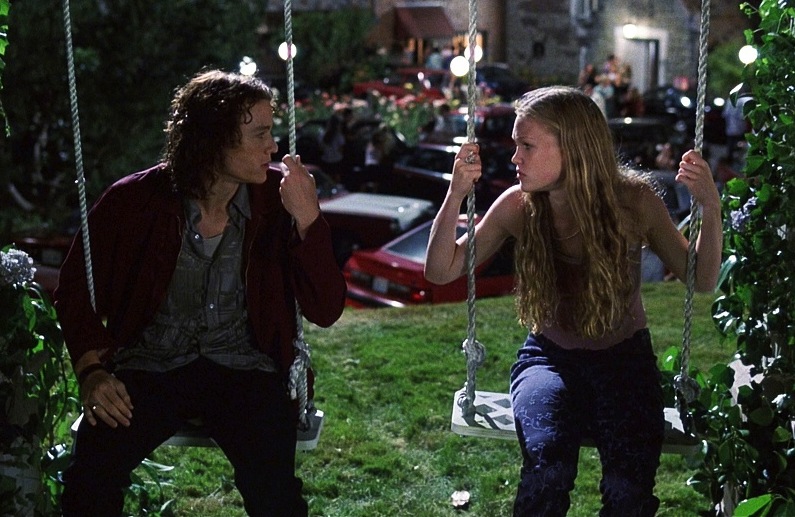}\\[1mm]
\begin{scriptsize}
\addtolength{\tabcolsep}{-2.1pt}
\begin{tabular}{m{0.00cm}m{3.63cm}m{0.00cm}m{4.1cm}m{0.00cm}m{3.72cm}m{0.02cm}m{4cm}}
 & \hspace{-3.6mm}{\color{cornellred}{\bf Q}}: How does E.T. show his happiness that he is finally returning home? & & \hspace{-3.6mm}{\color{cornellred}{\bf Q}}: Why do Joy and Jack get married that first night they meet in Las Vegas? & &  \hspace{-3.6mm}{\color{cornellred}{\bf Q}}: Why does Forrest undertake a three-year marathon? &  & \vspace{-2.6mm}\hspace{-3.6mm}{\color{cornellred}{\bf Q}}: How does Patrick start winning Kat over?\\[-1mm]
& \hspace{-3.4mm}{\color{cadmiumgreen}{\bf A}}: His heart lights up & & \hspace{-3.4mm}{\color{cadmiumgreen}{\bf A}}: They are both vulnerable and totally drunk & & \hspace{-3.4mm}{\color{cadmiumgreen}{\bf A}}: Because he is upset that Jenny left him & & \hspace{-3.4mm}{\color{cadmiumgreen}{\bf A}}: By getting personal information about her likes and dislikes
\end{tabular}
\end{scriptsize}
\vspace{-3mm}
\caption{\small Examples from the MovieQA dataset.
For illustration we show a single frame, however, all these questions/answers are time-stamped to a much longer clip in the movie.
Notice that while some questions can be answered using vision or dialogs alone, most require both.
Vision can be used to locate the scene set by the question, and semantics extracted from dialogs can be used to answer it.}
\label{fig:example_questions}
\vspace{-3mm}
\end{figure*}

Fast progress in Deep Learning as well as a large amount of available labeled data has significantly pushed forward the performance in many visual tasks such as image tagging, object detection and segmentation, action recognition, and image/video captioning.
We are steps closer to applications such as assistive solutions for the visually impaired, or cognitive robotics, which require a holistic understanding of the visual world by reasoning about all these tasks in a common framework.
However, a truly intelligent machine would ideally also infer high-level semantics underlying human actions such as motivation, intent and emotion, in order to react and, possibly, communicate appropriately.
These topics have only begun to be explored in the literature~\cite{thewhy,ZhuICCV15}.


A great way of showing one's understanding about the scene is to be able to answer any question about it~\cite{malinowski14nips}.
This idea gave rise to several question-answering datasets which provide a set of questions for each image along with multi-choice answers.
These datasets are either based on RGB-D images~\cite{malinowski14nips} or a large collection of static photos such as Microsoft COCO~\cite{VQA,VisualMadlibs}.
The types of questions typically asked are ``What'' is there and ``Where'' is it, what attributes an object has, what is its relation to other objects in the scene, and ``How many'' objects of certain type are present.
While these questions verify the holistic nature of our vision algorithms, there is an inherent limitation in what can be asked about a static image.
High-level semantics about actions and their intent is mostly lost and can typically only be inferred from temporal, possibly life-long visual observations.

Movies provide us with snapshots from people's lives that link into stories, allowing an experienced human viewer to get a high-level understanding of the characters, their actions, and the motivations behind them. 
Our goal is to create a question-answering dataset to evaluate machine comprehension of both, complex videos such as movies and their accompanying text.
We believe that this data will help push automatic semantic understanding to the next level, required to truly understand stories of such complexity.

This paper introduces MovieQA, a large-scale question-answering dataset about movies.
Our dataset consists of 14,944 multiple-choice questions with five deceiving options, of which only one is correct, sourced from 408 movies with high semantic diversity.
For 140 of these movies (6,462 QAs), we have timestamp annotations indicating the location of the question and answer in the video.
The questions range from simpler ``Who'' did ``What'' to ``Whom'' that can be solved by vision alone, to ``Why'' and ``How'' something happened, that can only be solved by exploiting both the visual information and dialogs (see Fig.~\ref{fig:example_questions} for a few example ``Why'' and ``How'' questions).
Our dataset is unique in that it contains multiple sources of information: video clips, subtitles, scripts, plots, and DVS~\cite{Rohrbach15} as illustrated in Fig.~\ref{fig:frontpage}.
We analyze the data through various statistics and intelligent baselines that mimic how different ``students'' would approach the quiz.
We further extend existing QA techniques to work with our data and show that question-answering with such open-ended semantics is hard.
We have created an online benchmark with a leaderboard (\url{http://movieqa.cs.toronto.edu/leaderboard}), encouraging inspiring work in this challenging domain.


\vspace{-1mm}
\section{Related work}
\label{sec:relwork}
\vspace{-1mm}

Integration of language and vision is a natural step towards improved understanding and is receiving increasing attention from the research community.
This is in large part due to efforts in large-scale data collection such as Microsoft's COCO~\cite{lin2014microsoft}, Flickr30K~\cite{Flickr30k} and Abstract Scenes~\cite{Abstract15} providing tens to hundreds of thousand images with natural language captions.
Having access to such data enabled the community to shift from hand-crafted language templates typically used for image description~\cite{BabyTalk} or retrieval-based approaches~\cite{Farhadi10,Im2Txt,Yang11} to deep neural models~\cite{Zitnick14,Karpathy15,kiros15,Vinyals14} that achieve impressive captioning results.
Another way of conveying semantic understanding of both vision and text is by retrieving semantically meaningful images given a natural language query~\cite{Karpathy15}.
An interesting direction, particularly for the goals of our paper, is also the task of learning common sense knowledge from captioned images~\cite{Vedantam15}.
This has so far been demonstrated only on synthetic clip-art scenes which enable perfect visual parsing.

{\bf Video understanding via language.}
In the video domain, there are fewer works on integrating vision and language, likely due to less available labeled data.
In~\cite{lrcn2014,Venugopalan:2014wc}, the authors caption video clips using LSTMs, ~\cite{rohrbach13iccv} formulates description as a machine translation model, while older work uses templates~\cite{SanjaUAI12,Das:2013br,krishnamoorthy:aaai13}.
In~\cite{Lin:2014db}, the authors retrieve relevant video clips for natural language queries, while~\cite{ramanathan13} exploits captioned clips to  learn action and role models.
For TV series in particular, the majority of work aims at recognizing and tracking characters in the videos~\cite{Baeuml2013_SemiPersonID,Bojanowski:2013bg,Ramanathan:2014fj,Sivic:2009kt}.
In~\cite{Cour08,Sankar:bv}, the authors aligned videos with movie scripts in order to improve scene prediction.
~\cite{Tapaswi_J1_PlotRetrieval} aligns movies with their plot synopses with the aim to allow semantic browsing of large video content via textual queries.
Just recently,~\cite{Tapaswi2015_Book2Movie,ZhuICCV15} aligned movies to books with the aim to ground temporal visual data with verbose and detailed descriptions available in books.

{\bf Question-answering.}
QA is a popular task in NLP with significant advances made recently with neural models such as memory networks~\cite{Sukhbaatar2015}, deep LSTMs~\cite{Hermann15}, and structured prediction~\cite{wang2015mctest}.
In computer vision,~\cite{malinowski14nips} proposed a Bayesian approach on top of a logic-based QA system~\cite{Liang13}, while~\cite{Malinowski15,mengye15} encoded both an image and the question using an LSTM and decoded an answer.
We are not aware of QA methods addressing the temporal domain.

{\bf QA Datasets.}
Most available datasets focus on image~\cite{KongCVPR14,lin2014microsoft,Flickr30k,Abstract15} or video description~\cite{Chen11,Rohrbach15,YouCook}.
Particularly relevant to our work is the MovieDescription dataset~\cite{Rohrbach15} which transcribed text from the Described Video Service (DVS), a narration service for the visually impaired, for a collection of over 100 movies.
For QA, \cite{malinowski14nips} provides questions and answers (mainly lists of objects, colors, \etc) for the NYUv2 RGB-D dataset, while~\cite{VQA,VisualMadlibs} do so for MS-COCO with a dataset of a million QAs.
While these datasets are unique in testing the vision algorithms in performing various tasks such as recognition, attribute induction and counting, they are inherently limited to static images.
In our work, we collect a large QA dataset sourced from over 400 movies with challenging questions that require semantic reasoning over a long temporal domain.

Our dataset is also related to purely text QA datasets such as MCTest~\cite{MCTest} which contains 660 short stories with 4 multi-choice QAs each, and~\cite{Hermann15} which converted 300K news summaries into Cloze-style questions.
We go beyond these datasets by having significantly longer text, as well as multiple sources of available information (plots, subtitles, scripts and DVS).
This makes our data one of a kind.

\vspace{-1.5mm}
\section{MovieQA dataset}
\label{sec:movieqa}
\vspace{-1mm}


\begin{table}[t]
 \vspace{-2.5mm}
\centering
\tabcolsep=0.19cm
{\small
\begin{tabular}{lrrrr}

\toprule
                       & \textsc{Train}   & \textsc{Val}    & \textsc{Test}   & \textsc{Total}               \\
\hline
\multicolumn{5}{>{\columncolor{bgr}}c}{Movies with Plots and Subtitles} \\[0.1mm]
\hline
 \#Movies              & 269     & 56     & 83     & 408                 \\
 \#QA                  & 9848    & 1958   & 3138   & 14944               \\
 Q \#words             & 9.3     & 9.3    & 9.5    & 9.3 $\pm$ 3.5       \\
 CA. \#words           & 5.7     & 5.4    & 5.4    & 5.6 $\pm$ 4.1       \\
 WA. \#words           & 5.2     & 5.0    & 5.1    & 5.1 $\pm$ 3.9       \\
\hline
\multicolumn{5}{>{\columncolor{orr}}c}{Movies with Video Clips} \\[0.1mm]
\hline
 \#Movies              & 93      & 21     & 26     & 140               \\
 \#QA                  & 4318    & 886    & 1258   & 6462              \\
 \#Video clips         & 4385    & 1098   & 1288   & 6771              \\
 Mean clip dur. (s)    & 201.0   & 198.5  & 211.4  & 202.7 $\pm$ 216.2 \\
 Mean QA \#shots       & 45.6    & 49.0   & 46.6   & 46.3  $\pm$ 57.1  \\
\bottomrule
\end{tabular}
}
\vspace*{-0.3cm}
\caption{MovieQA dataset stats.
Our dataset supports two modes of answering: text and video.
We present the split into train, val, and test splits for the number of movies and questions.
We also present mean counts with standard deviations in the total column.}
\vspace*{-0.35cm}
\label{tab:qa_stats}
\end{table}

The goal of our paper is to create a challenging benchmark that evaluates semantic understanding over long temporal data.
We collect a dataset with very diverse sources of information that can be exploited in this challenging domain.
Our data consists of quizzes about movies that the automatic systems will have to answer.
For each movie, a quiz comprises of a set of questions, each with 5 multiple-choice answers, only one of which is correct.
The system has access to various sources of textual and visual information, which we describe in detail below.

We collected 408 subtitled movies, and obtained their extended summaries in the form of plot synopses from \emph{Wikipedia}.
We crawled \emph{imsdb} for scripts, which were available for 49\% (199) of our movies.
A fraction of our movies (60) come with DVS transcriptions provided by~\cite{Rohrbach15}.

{\bf Plot synopses}
are movie summaries that fans write after watching the movie.
Synopses widely vary in detail and range from one to 20 paragraphs, but focus on describing content that is directly relevant to the story.
They rarely contain detailed visual information (\eg~character appearance), and focus more on describing the movie events and  character interactions.
We exploit 
plots to gather our quizzes.

{\bf Videos and subtitles.}
An average movie is about 2 hours in length and has over 198K frames and almost 2000 shots.
Note that video alone contains information about e.g., ``Who'' did ``What'' to ``Whom'', but may be lacking in information to explain why something happened.
Dialogs play an important role, and only both modalities together allow us to fully understand the story.
Note that subtitles do not contain speaker information. In our dataset, we provide video clips rather than full movies.

{\bf DVS} is a service that narrates movie scenes to the visually impaired by inserting relevant descriptions in between dialogs.
These descriptions contain sufficient ``visual'' information about the scene that they allow visually impaired audience to follow the movie.
DVS thus acts as a proxy for a perfect vision system, and is another source for answering.

{\bf Scripts.}
The scripts that we collected are written by screenwriters and serve as a guideline for movie making.
They typically contain detailed descriptions of scenes, and, unlike subtitles, contain both dialogs and speaker information.
Scripts are thus similar, if not richer in content to DVS+Subtitles, however are not always entirely faithful to the movie as the director may aspire to artistic freedom.

\vspace{-1mm}
\subsection{QA Collection method}
\vspace{-1mm}

\begin{table*}[t]
\centering
\tabcolsep=0.16cm
{\small
\begin{tabular}{lccclllrr}
\toprule
                                    & Txt     & Img     & Vid     & Goal
                                    & Data source       & AType     & \#Q       & AW \\
\midrule
MCTest~\cite{MCTest}                & \cmark  & -       & -       & reading comprehension
                                    & Children stories  & MC (4)    & 2,640     & 3.40  \\

bAbI~\cite{Weston14}                & \cmark  & -       & -       & reasoning for toy tasks
                                    & Synthetic         & Word      & 20$\times$2,000 & 1.0 \\

CNN+DailyMail~\cite{Hermann15}      & \cmark  & -       & -       & information abstraction
                                    & News articles     & Word      & 1,000,000* & 1*    \\

DAQUAR~\cite{malinowski14nips}      & -       & \cmark  & -       & visual: counts, colors, objects
                                    & NYU-RGBD          & Word/List & 12,468    & 1.15 \\

Visual Madlibs~\cite{VisualMadlibs} & -       & \cmark  & -       & visual: scene, objects, person, ...
                                    & COCO+Prompts      & FITB/MC (4)& 2$\times$75,208* & 2.59 \\

VQA (v1)~\cite{VQA}                 & -       & \cmark  & -       & visual understanding
                                    & COCO+Abstract     & Open/MC (18) & 764,163   & 1.24 \\

MovieQA                             & \cmark  & \cmark  & \cmark  & text+visual story comprehension
                                    & Movie stories     & MC (5)    & 14,944     & 5.29  \\
\bottomrule
\end{tabular}
}
\vspace{-0.25cm}
\caption{A comparison of various QA datasets. First three columns depict the modality in which the story is presented. 
AType: answer type; AW: average \# of words in answer(s); MC (N): multiple choice with N answers; FITB: fill in the blanks; *estimated information.}
\vspace{-0.3cm}
\label{tab:dataset-comparison}
\end{table*}

\begin{figure}
 \vspace{-2.5mm}
\centering
  \includegraphics[width=0.93\linewidth,trim=0 0 0 20,clip]{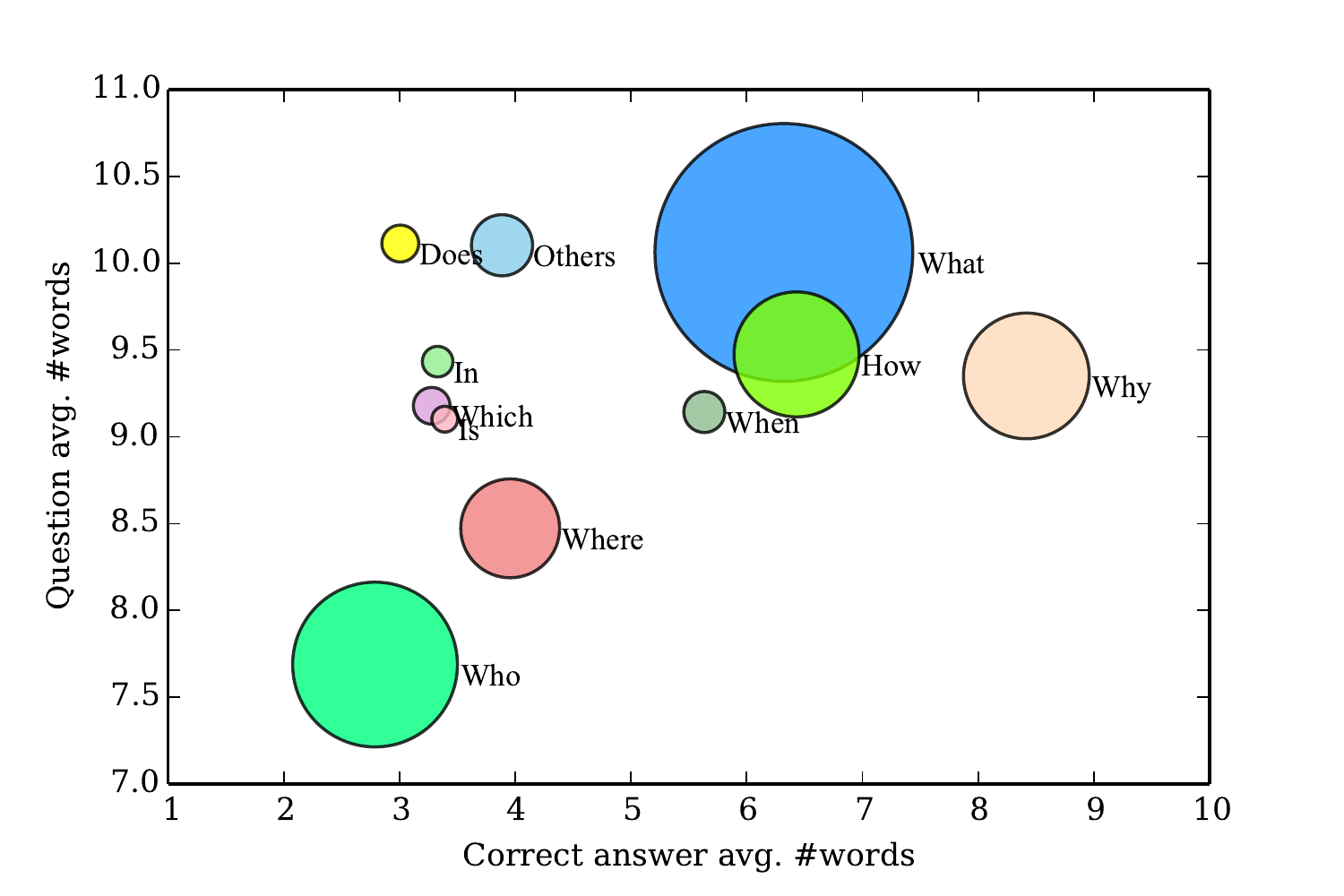}
\vspace*{-0.4cm}
\caption{Average number of words in MovieQA dataset based on the first word in the question. Area of a bubble indicates \#QA.}
\vspace*{-0.4cm}
\label{fig:stats:qword_calength}
\end{figure}

Since videos are difficult and expensive to provide to annotators, we used plot synopses as a proxy for the movie.
While creating quizzes, our annotators only referred to the story plot and were thus automatically coerced into asking story-like questions.
We split our annotation efforts into two primary parts to ensure high quality of the collected data.

{\bf Q and correct A.}
Our annotators were first asked to select a movie from a large list, and were shown its plot synopsis one paragraph at a time.
For each paragraph, the annotator had the freedom of forming any number and type of questions.
Each annotator was asked to provide the correct answer, and was additionally required to mark a minimal set of sentences within the plot synopsis paragraph that can be used to both frame the question and answer it.
This was treated as ground-truth for localizing the QA in the plot.

In our instructions, we asked the annotators to provide context to each question, such that a human taking the quiz should be able to answer it by watching the movie alone (without having access to the synopsis).
The purpose of this was to ensure questions that are localizable in the video and story as opposed to generic questions such as ``What are they talking?".
We trained our annotators for about one to two hours and gave them the option to re-visit and correct their data.
The annotators were paid by the hour, a strategy that allowed us to collect more thoughtful and complex QAs, rather than short questions and single-word answers.

{\bf Multiple answer choices.}
In the second step of data collection, we collected multiple-choice answers for each question.
Our annotators were shown a paragraph and a question at a time, but not the correct answer.
They were then asked to answer the question correctly as well as provide 4 wrong answers.
These answers were either deceiving facts from the same paragraph or common-sense answers.
The annotator was also allowed to re-formulate or correct the question.
We used this to sanity check all the questions received in the first step.
All QAs from the ``val'' and ``test'' set underwent another round of clean up.

{\bf Time-stamp to video.}
We further asked in-house annotators to align each sentence in the plot synopsis to the video by marking the beginning and end (in seconds) of the video that the sentence describes.
Long and complicated plot sentences were often aligned to multiple, non-consecutive video clips.
Annotation took roughly 2 hours per movie.
Since we have each QA aligned to a sentence(s) in the plot synopsis, the video to plot alignment links QAs with video clips.
We provide these clips as part of our benchmark.

\subsection{Dataset Statistics}

\begin{figure}
\vspace{-4mm}
\centering
  \includegraphics[width=0.9\linewidth]{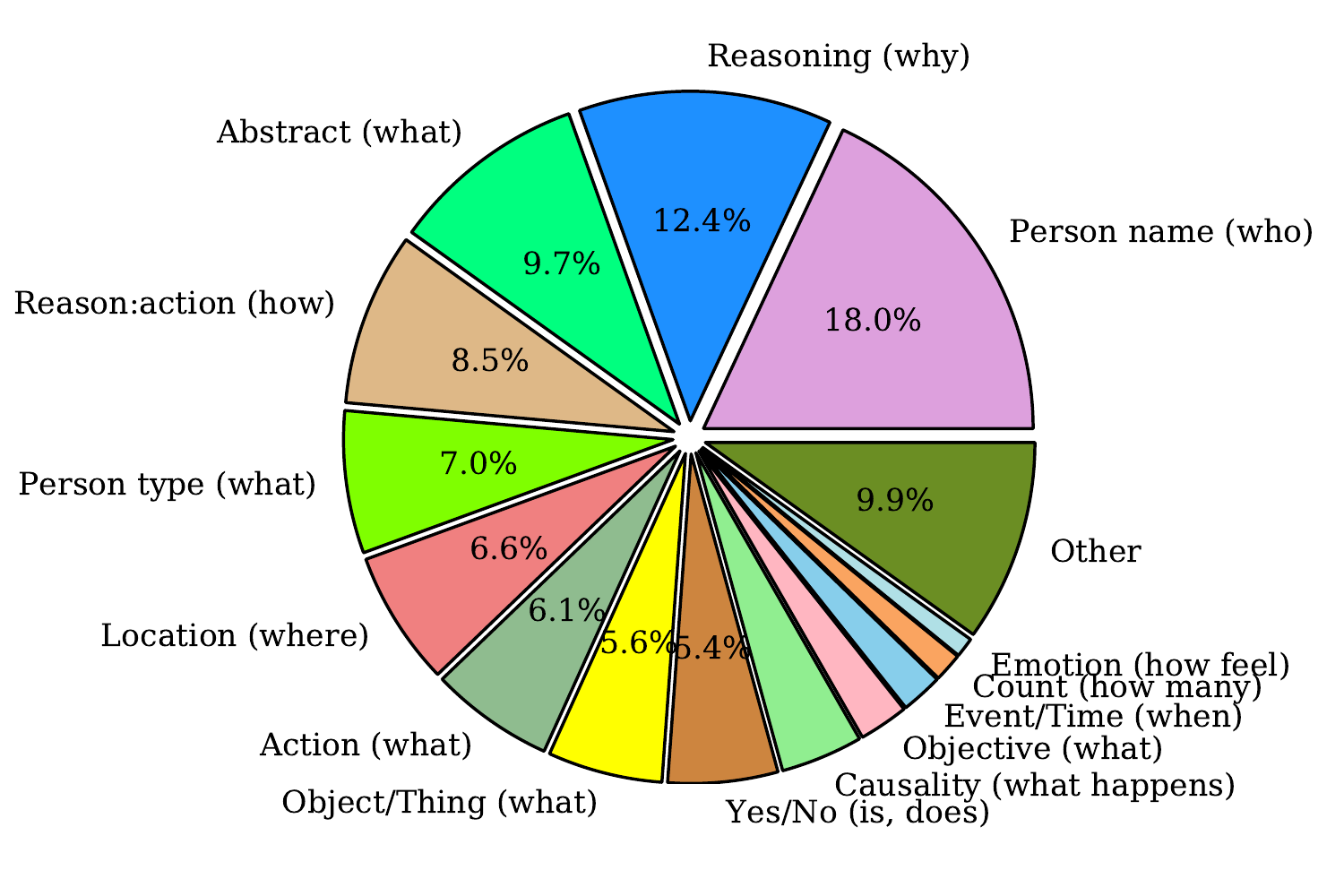}
\vspace*{-0.6cm}
\caption{Stats about MovieQA questions based on answer types.
Note how questions beginning with the same word may cover a variety of answer types:
\emph{Causality}: What happens ... ?; \emph{Action}: What did X do?
\emph{Person name}: What is the killer's name?; \etc 
}
\vspace*{-0.5cm}
\label{fig:stats:answer_type}
\end{figure}

In the following, we present some statistics of our MovieQA dataset.
Table~\ref{tab:dataset-comparison} presents an overview of popular and recent Question-Answering datasets in the field.
Most datasets (except MCTest) use very short answers and are thus limited to covering simpler visual/textual forms of understanding.
To the best of our knowledge, our dataset not only has long sentence-like answers, but is also the first to use videos in the form of movies.

{\bf Multi-choice QA.}
We collected a total of 14,944 QAs from 408 movies.
Each question comes with one correct and four deceiving answers.
Table~\ref{tab:qa_stats} presents an overview of the dataset along with information about the train/val/test splits, which will be used to evaluate automatically trained QA models.
On average, our questions and answers are fairly long with about 9 and 5 words respectively unlike most other QA datasets.
The video-based answering split for our dataset, supports 140 movies for which we aligned plot synopses with videos.
Note that the QA methods needs to look at a long video clip ($\sim$200s) to answer the question.

Fig.~\ref{fig:stats:qword_calength} presents the number of questions (bubble area) split based on the first word of the question along with information about number of words in the question and answer.
Of particular interest are ``Why'' questions that require verbose answers, justified by having the largest average number of words in the correct answer, and in contrast, ``Who'' questions with answers being short people names.

Instead of the first word in the question, a peculiar way to categorize QAs is based on the answer type.
We present such an analysis in Fig.~\ref{fig:stats:answer_type}.
Note how reasoning based questions (Why, How, Abstract) are a large part of our data.
In the bottom left quadrant we see typical question types that can likely be answered using vision alone.
Note however, that even the reasoning questions typically require vision, as the question context provides a visual description of a scene (\eg,~``Why does John run after Mary?'').


\begin{table}[t]
\centering
{\small
\begin{tabular}{lccc}
\toprule
 Text type   &   \# Movies &   \# Sent. / Mov. &  \# Words in Sent. \\
\midrule
 Plot        &     408 &        ~~~~35.2 &         20.3 \\
 Subtitle    &     408 &          1558.3 &        ~~6.2 \\
 Script      &     199 &          2876.8 &        ~~8.3 \\
 DVS         &    ~~60 &         ~~636.3 &        ~~9.3 \\
\bottomrule
\end{tabular}
}
\vspace*{-0.25cm}
\caption{Statistics for the various text sources used for answering.}
\vspace*{-0.4cm}
\label{tab:textsrc_stats}
\end{table}

{\bf Text sources for answering.}
In Table~\ref{tab:textsrc_stats}, we summarize and present some statistics about different text sources used for answering.
Note how plot synopses have a large number of words per sentence, hinting towards the richness and complexity of the source.


\section{Multi-choice Question-Answering}

We now investigate a number of intelligent baselines for QA.
We also study inherent biases in the data and try to answer the quizzes based simply on answer characteristics such as word length or within answer diversity.

Formally, let $S$ denote the story, which can take the form of any of the available sources of information -- \eg~plots, subtitles, or video shots.
Each story $S$ has a set of questions, and we assume that the (automatic) student reads one question $q^S$ at a time. 
Let $\{a_{j}^S\}_{j=1}^{M}$ be the set of multiple choice answers (only one of which is correct) corresponding to $q^S$, with $M=5$ in our dataset.

The general problem of multi-choice question answering can be formulated by a three-way scoring function $f(S,q^S,a^S)$. This function evaluates the ``quality'' of the answer given the story and the question. 
Our goal is thus to pick the best answer $a^S$ for question $q^S$ that maximizes $f$:
\begin{equation}
j^* = \arg\max_{j=1\ldots M} f(S, q^S, a_{j}^S) \, 
\end{equation}
Answering schemes are thus different functions $f$.
We drop the superscript $(\cdot)^S$ for simplicity of notation.

\subsection{The Hasty Student}
\label{sec:hasty_student}

We first consider $f$ which ignores the story and attempts to answer the question directly based on latent biases and similarities.
We call such a baseline as the ``Hasty Student'' since he/she is not concerned to read/watch the actual story.

The extreme case of a hasty student is to try and answer the question by only looking at the answers.
Here, $f(S, q, a_{j}) = g_{H1}(a_{j}|{\bf a})$, where $g_{H1}(\cdot)$ captures some properties of the answers.

{\bf Answer length.}
We explore using the number of words in the multiple choices to find the correct answer and explore biases in the dataset.
As shown in Table~\ref{tab:qa_stats}, correct answers are slightly longer as it is often difficult to frame long deceiving answers.
We choose an answer by:
(i) selecting the longest answer;
(ii) selecting the shortest answer; or
(iii) selecting the answer with the most different length.

{\bf Within answer similarity/difference.}
While still looking only at the answers, we compute a distance between all answers based on their representations (discussed in Sec.~\ref{sec:representation}).
We then select our answer as either the most similar or most distinct among all answers.

{\bf Q and A similarity.}
We now consider a hasty student that looks at both the question and answer,  $f(S, q, a_j) = g_{H2}(q, a_{j})$.
We compute similarity between the question and each answer and pick the highest scoring answer.

\subsection{The Searching Student}
\label{sec:sliding}

While the hasty student ignores the story, we consider a student that tries to answer the question by trying to locate a subset of the story $S$ which is most similar to both the question and the answer.
The scoring function $f$ is
\begin{equation}
f(S, q, a_{j}) = g_I(S, q) + g_I(S, a_{j}) \, .
\end{equation}
a factorization of the question and answer similarity.
We propose two similarity functions:
a simple windowed cosine similarity, and another using a neural architecture.

{\bf Cosine similarity with a sliding window.} We aim to find the best window of $H$ sentences (or shots) in the story $S$ that maximize similarity between the story and question, and story and answer.
We define our similarity function:
\begin{equation}
f(S, q, a_{j}) = \max_l \sum_{k = l}^{l+H} g_{ss}(s_k, q) + g_{ss}(s_k, a_{j}) \, ,
\end{equation}
where $s_k$ denotes a sentence (or shot) from the story $S$.
We use $g_{ss}(s, q) = x(s)^T x(q)$ as a dot product between the (normalized) representations of the two sentences (shots).
We discuss these representations in detail in Sec.~\ref{sec:representation}.

{\bf Searching student with a convolutional brain (SSCB).}
Instead of factoring $f(S, q, a_{j})$ as a fixed (unweighted) sum of two similarity functions $g_{I}(S, q)$ and $g_{I}(S, a_{j})$, we build a neural network that learns such a function.
Assuming the story $S$ is of length $n$, \eg~$n$ plot sentences or $n$ video shots, $g_{I}(S, q)$ and $g_{I}(S, a_{j})$ can be seen as two vectors of length $n$ whose $k$-th entry is $g_{ss}(s_k, q)$.
We further combine all $[g_I(S, a_{j})]_j$ for the 5 answers into a $n\times 5$ matrix.
The vector $g_{I}(S, q)$ is replicated $5$-times, and we stack the question and answer matrix together to obtain a tensor of size $n \times 5 \times 2$.

\begin{figure}
\vspace{-5mm}
\centering
  \includegraphics[width=0.95\linewidth,trim=0 0 0 0,clip]{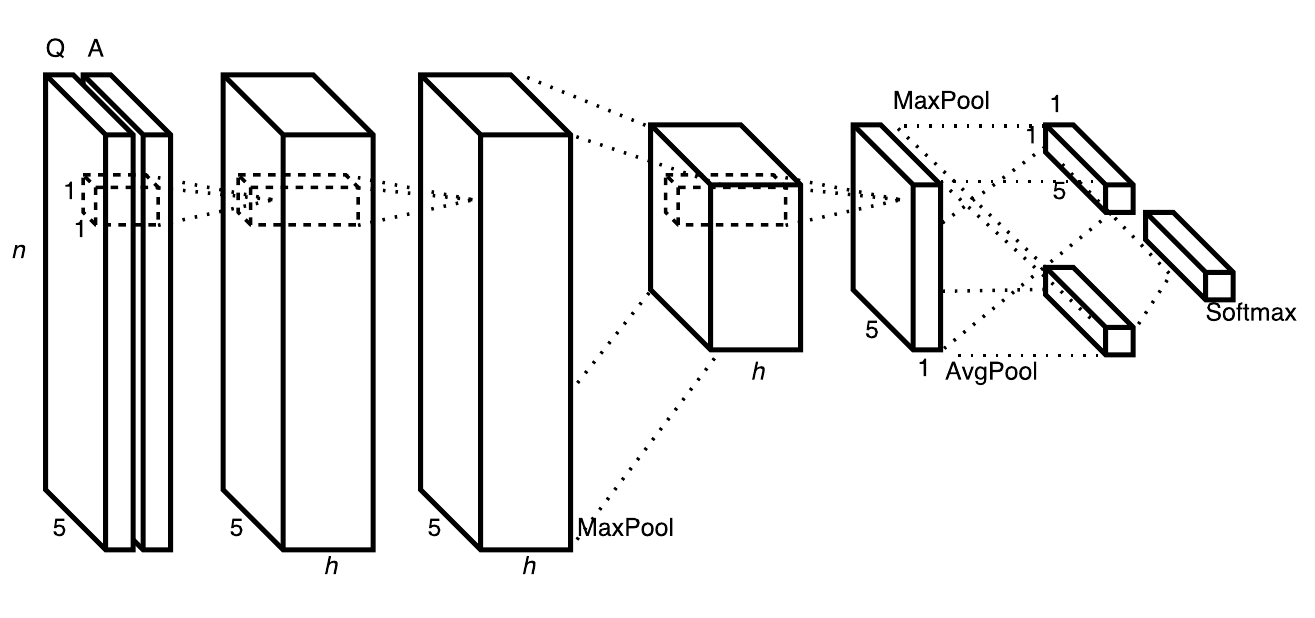}
\vspace*{-0.5cm}
\caption{\small Our neural similarity architecture (see text for details).}
\label{fig:model:cnn}
\vspace*{-0.5cm}
\end{figure}

Our neural similarity model is a convnet (CNN), shown in Fig.~\ref{fig:model:cnn}, that takes the above tensor, and applies couple layers of $h = 10$, $1 \times 1$ convolutions to approximate a family of functions $\phi(g_I(S, q), g_I(S, a_{j}))$.
Additionally, we incorporate a max pooling layer with kernel size $3$ to allow for scoring the similarity within a window in the story.
The last convolutional output is a tensor with shape ($\frac{n}{3}, 5$), and we apply both mean and max pooling across the storyline, add them, and make predictions using softmax.
We train our network using cross-entropy loss and the Adam optimizer~\cite{kingma2014adam}.

\subsection{Memory Network for Complex QA}

Memory Networks were originally proposed for text QA and model complex three-way relationships between the story, question and answer.
We briefly describe MemN2N proposed by~\cite{Sukhbaatar2015} and suggest simple extensions to make it suitable for our data and task.

The input of the original MemN2N is a story and question.
The answering is restricted to single words and is done by picking the most likely word from the vocabulary $\mathcal{V}$ of 20-40 words.
Note that this is not directly applicable to MovieQA, as our data set does not have perform vocabulary-based answering.

A question $q$ is encoded as a vector $u \in \mathbb{R}^d$ using a word embedding $B \in \mathbb{R}^{d \times |\mathcal{V}|}$.
Here, $d$ is the embedding dimension, and $u$ is obtained by mean-pooling the representations of words in the question.
Simultaneously, the sentences of the story $s_l$ are encoded using word embeddings $A$ and $C$ to provide two different sentence representations $m_l$ and $c_l$, respectively.
$m_l$, the representation of sentence $l$ in the story, is used in conjunction with $u$ to produce an attention-like mechanism which selects sentences in the story most similar to the question via a softmax function:
\begin{equation}
p_l = \mathrm{softmax}(u^T m_l) \, .
\end{equation}
The probability $p_l$ is used to weight the second sentence embedding $c_l$, and the output $o = \sum_l p_l c_l$ is obtained by pooling the weighted sentence representations across the story.
Finally, a linear projection $W \in \mathbb{R}^{|\mathcal{V}| \times d}$ decodes the question $u$ and the story representation $o$ to provide a soft score for each vocabulary word
\begin{equation}
a = \mathrm{softmax}(W (o + u)) \, .
\end{equation}
The top scoring word $\hat a$ is picked from $a$ as the answer.
The free parameters to train are the embeddings $B$, $A$, $C$, $W$ for different words which can be shared across different layers.

Due to its fixed set of output answers, the MemN2N in the current form is not designed for multi-choice answering with open, natural language answers.
We propose two key modifications to make the network suitable for our task.

{\bf MemN2N for natural language answers.}
To allow the MemN2N to rank multiple answers written in natural language, we add an additional embedding layer $F$ which maps each multi-choice answer $a_j$ to a vector $g_j$.
Note that $F$ is similar to embeddings $B$, $A$ and $C$, but operates on answers instead of the question or story.
To predict the correct answer, we compute the similarity between the answers $g$, the question embedding $u$ and the story representation $o$:
\begin{equation}
\label{eq:memnet_multichoice_ans}
a = \mathrm{softmax}((o + u)^T g)
\end{equation}
and pick the most probable answer as correct.
In our general QA formulation, this is equivalent to
\begin{equation}
f(S, q, a_{j}) = g_{M1}(S, q, a_{j}) + g_{M2}(q, a_{j}),
\end{equation}
where $g_{M1}$ attends to parts of the story using the question, and a second function $g_{M2}$ directly considers similarities between the question and the answer.

{\bf Weight sharing and fixed word embeddings.}
The original MemN2N learns embeddings for each word based directly on the task of question-answering.
However, to scale this to large vocabulary data sets like ours, this requires unreasonable amounts of training data.
For example, training a model with a vocabulary size 14,000 (obtained just from plot synopses) and $d = 100$ would entail learning 1.4M parameters for each embedding.
To prevent overfitting, we first share all word embeddings $B, A, C, F$ of the memory network.
Nevertheless, even one embedding is still a large number of parameters.

We make the following crucial modification that allows us to use the Memory Network for our dataset.
We drop $B$, $A$, $C$, $F$ and replace them by a fixed (pre-trained) word embedding $Z \in \mathbb{R}^{d_1 \times |\mathcal{V}|}$ obtained from the Word2Vec model and learn a shared linear projection layer $T \in \mathbb{R}^{d_2 \times d_1}$ to map all sentences (stories, questions and answers) into a common space.
Here, $d_1$ is the dimension of the Word2Vec embedding, and $d_2$ is the projection dimension.
Thus, the new encodings are
\begin{equation}
u = T \cdot Z q; \, m_l, c_l = T \cdot Z s_l; \, \mathrm{and} \, g_j = T \cdot Z a_j .
\end{equation}
Answer prediction is performed as before in Eq.~\ref{eq:memnet_multichoice_ans}.

We initialize $T$ either using an identity matrix $d_1 \times d_1$ or using PCA to lower the dimension from $d_1 = 300$ to $d_2 = 100$. 
Training is performed using stochastic gradient descent with a batch size of 32.

\subsection{Representations for Text and Video}
\label{sec:representation}

{\bf TF-IDF} is a popular and successful feature in information retrieval.
In our case, we treat plots (or other forms of text) from different movies as documents and compute a weight for each word.
We set all words to lower case, use stemming, and compute the vocabulary $\mathcal{V}$ which consists of words $w$ that appear more than $\theta$ times in the documents. 
We represent each sentence (or question or answer) in a bag-of-words style with an TF-IDF score for each word.

{\bf Word2Vec.}
A disadvantage of TF-IDF is that it is unable to capture the similarities between words.
We use the skip-gram model proposed by~\cite{mikolov2013efficient} and train it on roughly 1200 movie plots to obtain domain-specific, $300$ dimensional word embeddings.
A sentence is then represented by mean-pooling its word embeddings.
We normalize the resulting vector to have unit norm.

{\bf SkipThoughts.}
While the sentence representation using mean pooled Word2Vec discards word order, SkipThoughts~\cite{skipthoughts} use a Recurrent Neural Network to capture the underlying sentence semantics.
We use the pre-trained model by~\cite{skipthoughts} to compute a $4800$ dimensional sentence representation. 

{\bf Video.} 
To answer questions from the video, we learn an embedding between a shot and a sentence, which maps the two modalities in a common space. In this joint space, one can score the similarity between the two modalities via a simple dot product. This allows us to apply all of our proposed question-answering techniques in their original form.

To learn the joint embedding we follow~\cite{ZhuICCV15} which extends~\cite{kiros15} to video.
Specifically, we use the GoogLeNet architecture~\cite{szegedy2014going} as well as hybrid-CNN~\cite{ZhouNIPS2014} to extract frame-wise features, and mean-pool the representations over all frames in a shot.
The embedding is a linear mapping of the shot representation and an LSTM on word embeddings on the sentence side, trained using the ranking loss on the MovieDescription Dataset~\cite{Rohrbach15} as in~\cite{ZhuICCV15}.


\section{QA Evaluation}
\label{sec:eval}

We present results for question-answering with the proposed methods on our MovieQA dataset.
We study how various sources of information influence the performance, and how different levels of complexity encoded in $f$ affects the quality of automatic QA.

{\bf Protocol.}
Note that we have two primary tasks for evaluation.
(i) {\bf Text-based}: the story takes the form of various texts -- plots, subtitles, scripts, DVS; and
(ii) {\bf Video-based}: story is the video, and with/without subtitles.

{\bf Dataset structure.}
The dataset is divided into three disjoint splits: \emph{train}, \emph{val}, and \emph{test}, based on unique movie titles in each split.
The splits are optimized to preserve the ratios between \#movies, \#QAs, and all the story sources at 10:2:3 (\eg~about 10k, 2k, and 3k QAs).
Stats for each split are presented in Table~\ref{tab:qa_stats}.
The \emph{train} set is to be used for training automatic models and tuning any hyperparameters.
The \emph{val} set should not be touched during training, and may be used to report results for several models.
The \emph{test} set is a held-out set, and is evaluated on our MovieQA server.
For this paper, all results are presented on the \emph{val} set.

{\bf Metrics.}
Multiple choice QA leads to a simple and objective evaluation.
We measure \emph{accuracy}, the number of correctly answered QAs over the total count.

\vspace{-1mm}
\subsection{The Hasty Student}
\vspace{-1mm}
\begin{table}[t]
\centering
{\small
\begin{tabular}{l|lrrr}
\toprule
\multirow{2}{*}{{\bf Answer length}}     &            & longest & shortest & different \\
    &             & 25.33   & 14.56    & 20.38     \\
\midrule
\multirow{3}{*}{{\bf Within answers}} &                 & TF-IDF   & SkipT    & w2v       \\
 & similar         & 21.71   & 28.14    & 25.43     \\
 & distinct        & 19.92   & 14.91    & 15.12     \\
\midrule
\multirow{2}{*}{{\bf Question-answer}} &              & TF-IDF   & SkipT    & w2v       \\
 & similar         & 12.97   & 19.25    & 24.97     \\
\bottomrule
\end{tabular}
}
\vspace*{-0.2cm}
\caption{The question-answering accuracy for the ``Hasty Student'' who tries to answer questions without looking at the story.}
\vspace*{-0.5cm}
\label{tab:results-cheating_baseline}
\end{table}

The first part of Table~\ref{tab:results-cheating_baseline} shows the performance of three models when trying to answer questions based on the answer length.
Notably, always choosing the longest answer performs better (25.3\%) than random (20\%).
The second part of Table~\ref{tab:results-cheating_baseline} presents results when using within-answer feature-based similarity.
We see that the answer most similar to others is likely to be correct when the representations are generic and try to capture the semantics of the sentence (Word2Vec, SkipThoughts).
The most distinct answers performs worse than random on all features.
In the last section of Table~\ref{tab:results-cheating_baseline} we see that computing feature-based similarity between questions and answers is insufficient for answering.
Especially, TF-IDF performs worse than random since words in the question rarely appear in the answer.

{\bf Hasty Turker.}
To analyze the deceiving nature of our multi-choice QAs, we tested humans (via AMT) on a subset of 200 QAs.
The turkers were not shown the story in any form and were asked to pick the best possible answer given the question and a set of options.
We asked each question to 10 turkers, and rewarded each with a bonus if their answer agreed with the majority.
We observe that without access to the story, humans obtain an accuracy of 27.6\%.
We suspect that the bias is due to the fact that some of the QAs reveal the movie (e.g., ``Darth Vader'') and the turker may have seen this movie.
Removing such questions, and re-evaluating on a subset of 135 QAs, lowers the performance to 24.7\%.
This shows the genuine difficulty of our QAs.

\vspace{-1mm}
\subsection{Searching Student}
\vspace{-1mm}

{\bf Cosine similarity with window.}
The first section of Table~\ref{tab:results-comprehensive_baseline} presents results for the proposed cosine similarity using different representations and text stories.
Using the plots to answer questions outperforms other sources (subtitles, scripts, and DVS) as the QAs were collected using plots and annotators often reproduce words from the plot.

\begin{figure}
\centering
  \includegraphics[width=0.8\linewidth,trim=0 0 0 12,clip]{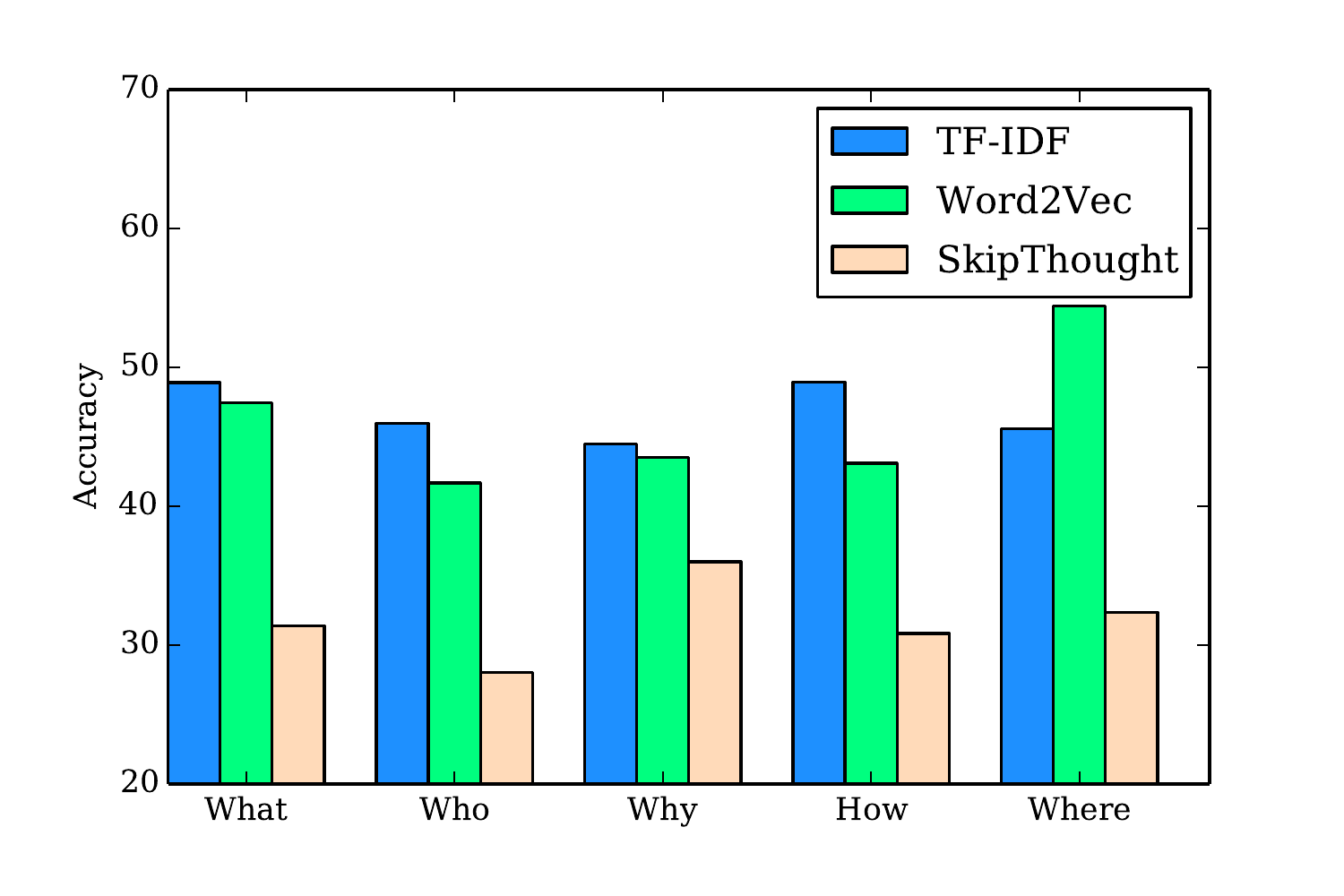}
\vspace*{-0.6cm}
\caption{Accuracy for different feature representations of plot sentences with respect to the first word of the question.}
\label{fig:results-simple_baselines_qfw}
\vspace{-0.5cm}
\end{figure}

We show the results of using Word2Vec or SkipThought representations in the following rows of Table~\ref{tab:results-comprehensive_baseline}.
SkipThoughts perform much worse than both TF-IDF and Word2Vec which are closer.
We suspect that while SkipThoughts are good at capturing the overall semantics of a sentence, proper nouns -- names, places -- are often hard to distinguish.
Fig.~\ref{fig:results-simple_baselines_qfw} presents a accuracy breakup based on the first word of the questions.
TF-IDF and Word2Vec perform considerably well, however, we see a larger difference between the two for ``Who'' and ``Why'' questions.
``Who'' questions require distinguishing between names, and ``Why'' answers are typically long, and mean pooling destroys semantics.
In fact Word2Vec performs best on ``Where'' questions that may use synonyms to indicate places.
SkipThoughts perform best on ``Why'' questions where sentence semantics help improve answering.

\begin{table}[t]
\vspace{-2.0mm}
\tabcolsep=0.20cm
\centering
{\small
\begin{tabular}{lcccc}
\toprule
Method      				&   Plot    &  DVS     & Subtitle &  Script \\
\midrule
Cosine TFIDF                &    47.6   &   24.5   &   24.5   &   24.6  \\
Cosine SkipThought          &    31.0   &   19.9   &   21.3   &   21.2  \\
Cosine Word2Vec             &    46.4   &   26.6   &   24.5   &   23.4  \\
\midrule
SSCB TFIDF                  &    48.5   &   24.5   &   27.6   &   26.1  \\
SSCB SkipThought            &    28.3   &   24.5   &   20.8   &   21.0  \\
SSCB Word2Vec               &    45.1   &   24.8   &   24.8   &   25.0  \\
\midrule
SSCB Fusion                 & {\bf56.7} &   24.8   &   27.7   &   28.7  \\
\midrule
MemN2N (w2v, linproj)       &    40.6   & {\bf33.0}& {\bf38.0}& {\bf42.3}\\
\bottomrule
\end{tabular}
}
\vspace*{-0.25cm}
\caption{Accuracy for Text-based QA. {\bf Top}: results for the Searching student with cosine similarity; {\bf Middle}: Convnet SSCB; and {\bf Bottom}: the modified Memory Network.}
\label{tab:results-comprehensive_baseline}
\vspace{-5.0mm}
\end{table}











{\bf SSCB}. The middle rows of Table~\ref{tab:results-comprehensive_baseline} show the result of our neural similarity model.
Here, we present additional results combining all text representations (\textit{SSCB fusion}) via our CNN.
We split the \emph{train} set into $90\%$ train / $10\%$ dev, such that all questions and answers of the same movie are in the same split, train our model on train and monitor performance on dev.
Both \emph{val} and \emph{test} sets are held out.
During training, we also create several model replicas and pick the ones with the best validation performance.

Table~\ref{tab:results-comprehensive_baseline} shows that the neural model outperforms the simple cosine similarity on most tasks, while the fusion method achieves the highest performance when using plot synopses as the story.
Ignoring the case of plots, the accuracy is capped at about $30\%$ for most modalities showing the difficulty of our dataset.

\vspace{-1mm}
\subsection{Memory Network}
\vspace{-0.5mm}

The original MemN2N which trains the word embeddings along with the answering modules overfits heavily on our dataset leading to near random performance on \textit{val} ($\sim$20\%).
However, our modifications help in restraining the learning process.
Table~\ref{tab:results-comprehensive_baseline} (bottom) presents results for MemN2N with Word2Vec initialization and a linear projection layer.
Using plot synopses, we see a performance closer to SSCB with Word2Vec features.
However, in the case of longer stories, the attention mechanism in the network is able to sift through thousands of story sentences and performs well on DVS, subtitles and scripts.
This shows that complex three-way scoring functions are needed to tackle such QA sources. 
In terms of story sources, the MemN2N performs best with scripts which contain the most information (descriptions, dialogs and speaker information).

\begin{table}[t]
\vspace{-2mm}
\centering
{\small
\begin{tabular}{lccc}

\toprule
Method                      &   Video   &  Subtitle    &  Video+Subtitle \\
\midrule
SSCB all clips              &  21.6     &  22.3        &    21.9      \\
MemN2N all clips            & \bf{23.1} &  \bf{38.0}   &   \bf{34.2}  \\
\bottomrule
\end{tabular}
}
\vspace*{-0.25cm}
\caption{Accuracy for Video-based QA and late fusion of Subtitle and Video scores.}
\label{tab:results-video_baseline}
\vspace{-3mm}
\end{table}



\vspace{-1mm}
\subsection{Video baselines}
\vspace{-1mm}
We evaluate SSCB and MemN2N in a setting where the automatic models answer questions by ``watching'' all the video clips that are provided for that movie.
Here, the story descriptors are shot embeddings.

The results are presented in Table~\ref{tab:results-video_baseline}.
We see that learning to answer questions using video is still a hard problem with performance close to random.
As visual information alone is insufficient, we also perform and experiment combining video and dialog (subtitles) through late fusion.
We train the SSCB model with the visual-text embedding for subtitles and see that it yields poor performance (22.3\%) compared to the fusion of all text features (27.7\%).
For the memory network, we answer subtitles as before using Word2Vec.


\vspace{-1mm}
\section{Conclusion}
\vspace{-1mm}
We introduced the MovieQA data set which aims to evaluate automatic story comprehension from both video and text. 
Our dataset is unique in that it contains several sources of information -- video clips, subtitles, scripts, plots and DVS. We provided several intelligent baselines and extended existing QA techniques to analyze the difficulty of our task. 
Our benchmark with an evaluation server is online at~\url{http://movieqa.cs.toronto.edu}.

{\small
\vspace{0.2cm}
\noindent {\bf Acknowledgment.}
We thank the \texttt{Upwork} annotators, Lea Jensterle, Marko Boben, and So\v{c}a Fidler for data collection, and Relu Patrascu for infrastructure support.
MT and RS are supported by DFG contract no. STI-598/2-1, and the work was carried out during MT's visit to U. of T. on a KHYS Research Travel Grant.
}

\balance
{\small
\bibliographystyle{ieee}
\bibliography{cvpr2016}
}

\end{document}